\documentclass[11pt,a4paper]{article}

\usepackage{CJKutf8}
\usepackage[utf8]{inputenc}
\usepackage[T1]{fontenc}

\usepackage[english]{babel}
\usepackage{amsmath}
\usepackage{amsfonts}
\usepackage{amssymb}
\usepackage{graphicx}
\usepackage{url}
\usepackage[sort]{natbib}
\usepackage{authblk}
\usepackage{geometry}
\usepackage{hyperref}
\usepackage{booktabs}
\usepackage{float}

\title{Japanese AI Agent System on Human Papillomavirus Vaccination: System Design}

\author[1]{Junyu Liu}
\author[2]{Siwen Yang}
\author[3]{Dexiu Ma}
\author[4]{Qian Niu}
\author[5]{Zequn Zhang}
\author[1]{\\Momoko Nagai-Tanima}
\author[1]{Tomoki Aoyama\thanks{Corresponding author: aoyama.tomoki.4e@kyoto-u.ac.jp}}

\affil[1]{Kyoto University, Kyoto, Japan}
\affil[2]{University of Waterloo, Waterloo, Canada}
\affil[3]{Texas Tech University, Lubbock, USA}
\affil[4]{The University of Tokyo, Tokyo, Japan}
\affil[5]{USTC, Hefei, China}

\date{}

\begin{document}

\maketitle

\section*{Abstract}

\textbf{Background:} Human papillomavirus (HPV) vaccine hesitancy poses significant public health challenges, particularly in Japan where proactive vaccination recommendations were suspended from 2013 to 2021. The resulting information gap between medical institutions and vaccine-hesitant populations is exacerbated by misinformation on social media. Traditional public health communication strategies cannot simultaneously address individual queries while monitoring population-level discourse.

\noindent\textbf{Objectives:} This study aimed to develop and evaluate a dual-purpose artificial intelligence agent system that provides verified HPV vaccine information to the public through a conversational interface while generating analytical reports for medical institutions based on user interactions and social media discourse.

\noindent\textbf{Methods:} We implemented a system comprising three components: a vector database integrating documents from academic papers, Japanese government sources, news media, and social media posts; a Retrieval-Augmented Generation chatbot using a ReAct agent architecture with iterative multi-tool orchestration across five specialized knowledge sources; and an automated report generation system with modules for news analysis, research synthesis, social media sentiment analysis including stance classification and topic modeling, and user interaction pattern identification. System performance was assessed through both automated and manual evaluation protocols using a 0-5 scoring scale.

\noindent\textbf{Results:} The whole system functions as expected. For single-turn evaluation, the chatbot achieved mean scores (SD) of 4.83 (0.67) for relevance, 4.89 (0.53) for routing, 4.50 (1.29) for reference quality, 4.90 (0.62) for correctness, and 4.88 (0.54) for professional identity, with an overall mean of 4.80 (0.80). Multi-turn evaluation yielded higher scores: context retention 4.94 (0.25), topic coherence 5.00 (0.00), and overall mean 4.98 (0.15), with topic-centering and identity achieving 5.00. The report generation system achieved high scores across all sections: completeness ranged from 4.00 to 5.00 (paper sections consistently perfect), correctness from 4.00 to 5.00, and helpfulness from 3.67 to 5.00. Reference validity achieved perfect scores (5.00) across all periods, with citation correctness averaging 4.33 (0.50) for news sections and 4.08 (0.66) for paper sections.

\noindent\textbf{Conclusions:} This study demonstrates the feasibility of an integrated AI agent system for bidirectional HPV vaccine communication in Japan. The architecture enables verified information delivery with source attribution while providing institutional stakeholders with systematic public discourse analysis. The transferable framework provides foundations for adaptation to other vaccines and multilingual public health contexts.
\newline
\newline
\noindent\textbf{Keywords:} HPV; AI agent; large language model; stance analysis; topic modeling

\section{Introduction}

Human papillomavirus (HPV) is a significant public health concern, causing approximately 662,044 new cases of cervical cancer and 348,709 deaths annually worldwide in 2022 \cite{wu2025global}. HPV vaccines have demonstrated high efficacy in preventing HPV-related diseases \cite{hpv_vaccine_efficacy}, with numerous countries implementing national vaccination programs since the vaccines' introduction in 2006. However, vaccine hesitancy remains a persistent challenge \cite{vaccine_hesitancy}, particularly in countries like Japan where HPV vaccination rates dropped dramatically following safety concerns and media coverage of alleged adverse events \cite{japan_hpv_suspension}.

The spread of misinformation about HPV vaccines through social media platforms has exacerbated public concerns \cite{vaccine_misinformation_social_media}, creating a complex information landscape where accurate medical information competes with anecdotal reports and unverified claims. Traditional public health communication strategies face significant challenges in addressing health misinformation at scale, as responding effectively requires simultaneously countering individual-level psychological barriers while monitoring population-level misinformation dynamics across diverse platforms. \cite{sylvia2020we}. Medical institutions require timely insights into public discourse to develop effective communication strategies, yet manual analysis of vast amounts of social media data and public inquiries is resource-intensive and time-consuming.

Recent advances in large language models (LLMs) and Retrieval-Augmented Generation (RAG) systems offer promising solutions for bridging this information gap \cite{rag_survey}. LLMs demonstrate remarkable capabilities in natural language understanding and generation across multiple languages \cite{llm_healthcare}, including Japanese, which presents unique challenges due to its complex writing system and grammatical structure \cite{japanese_nlp}. RAG systems combine the generative capabilities of LLMs with retrieval from curated knowledge bases, enabling responses grounded in verified information sources while maintaining conversational fluency.

Previous work has applied natural language processing to HPV-related social media analysis, primarily focusing on sentiment analysis and topic modeling \cite{twitter_vaccine_sentiment}. However, these approaches typically operate as passive analytical tools rather than active information dissemination systems. Chatbot systems for health information have been developed for various domains \cite{health_chatbots}, but few integrate multi-source retrieval spanning academic literature, official guidelines, news media, and social media discourse while simultaneously providing bidirectional communication between the public and health institutions.

The Japanese context presents unique challenges and opportunities for such a system. Japan experienced a dramatic suspension of proactive HPV vaccination recommendations from 2013 to 2021 due to safety concerns, resulting in vaccination rates falling below 1\% and creating a substantial gap in population immunity \cite{japan_hpv_suspension}. The government's 2022 resumption of vaccination recommendations necessitates renewed public education efforts \cite{ujiie2022resumption}. Furthermore, Japanese-language health information systems face technical challenges including multi-script processing (hiragana, katakana, kanji), medical terminology localization, and culturally appropriate communication styles.

In this study, we developed and implemented a comprehensive AI agent system designed to address both public information needs and institutional monitoring requirements for HPV vaccine discourse in Japan. Our system provides two main functions: (1) a RAG-based chatbot that answers public queries by retrieving and synthesizing information from academic papers, official documents, news articles, and social media posts; and (2) an analytics dashboard that generates reports for medical institutions based on aggregated chat histories and social media data. The system employs multi-source data collection, semantic search with vector embeddings, intelligent query routing, and automated evaluation frameworks.

\section{Methods}

\subsection{System Architecture}

We developed a multi-component AI agent system for HPV vaccine information dissemination and public opinion analysis. The system comprises three main modules: a multi-source data collection and storage system, a ReAct agent-based chatbot \cite{yao2022react} for public information queries, and a report generation system for medical institutions.

\begin{figure}[H]
\centering
\includegraphics[width=0.9\textwidth]{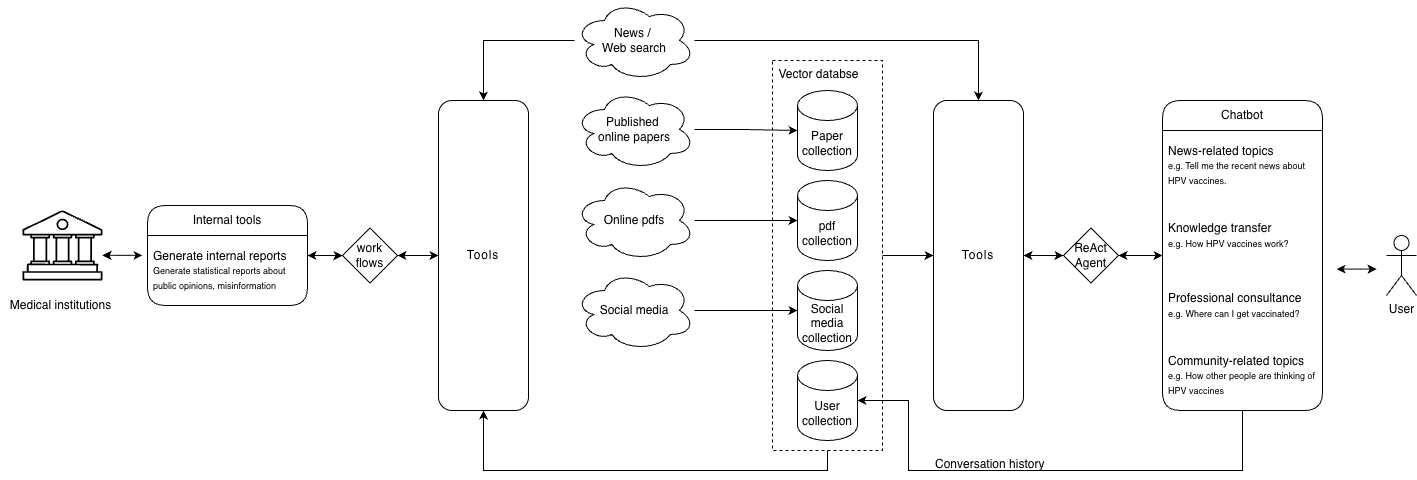}
\caption{Overall system architecture showing the integration of data collection, vector database, chatbot interface, and report generation components.}
\label{fig:system_architecture}
\end{figure}

The overall architecture follows a distributed design pattern with a centralized vector database (Qdrant) \cite{qdrant} serving as the knowledge repository (Figure \ref{fig:system_architecture}). Data flows from multiple external sources through specialized collectors into the database, where it is indexed using semantic embeddings. The chatbot and report generation modules both query this database but serve different end users with distinct interfaces and functionalities. The system implements a bidirectional information flow: the chatbot provides HPV vaccine information to the public while simultaneously collecting user inquiries with consent, and the report generator aggregates these interactions with social media data to produce actionable insights for medical institutions.

\subsection{Data Collection and Database}

We implemented a vector database infrastructure as the central knowledge repository, managing four distinct collections: academic papers, official documents, social media posts, and chat histories. Each document is represented as a 2048-dimensional vector using embedding models optimized for Japanese language processing (pfnet/plamo-embedding-1b) \cite{plamo_embedding}. The database employs cosine similarity metrics for semantic search operations \cite{sentence_bert,faiss_vector_search}, supporting efficient retrieval with customizable parameters and metadata preservation.

Data was collected from four heterogeneous sources to construct a comprehensive knowledge base spanning scientific evidence, official guidance, media coverage, and public discourse. Academic papers were retrieved from PubMed \cite{pubmed} through keyword-based searches with temporal filtering, capturing abstracts, medical subject headings terms, journal information, and DOI identifiers. Official documents and web content were collected from authoritative sources including the World Health Organization (WHO) \cite{who} and Japanese Ministry of Health, Labour and Welfare (MHLW) \cite{mhlw} through multiple complementary methods: intelligent query analysis for information synthesis, filtered web searches targeting official sources, online PDF document discovery and extraction, and specialized scraping of government meeting records and reference materials. News articles were aggregated from multiple news sources with keyword-based searches in both Japanese and English, employing deduplication to ensure unique coverage. Social media data from Twitter/X was collected through daily automated harvesting using Tweepy \cite{tweepy} with temporal specifications, implementing rate limit handling to ensure comprehensive data capture across extended time periods.

\subsection{Chatbot Implementation}

We implemented a ReAct agent-based chatbot using LlamaIndex \cite{llamaindex} employing an iterative multi-tool orchestration architecture where a single intelligent controller dynamically selects and combines information from multiple specialized data sources across sequential decision-making iterations. The system addresses the challenge of answering diverse user queries by enabling flexible, multi-source information gathering while maintaining conversational coherence and citation quality assurance.

\subsubsection{Architecture}

The chatbot employs a single controller agent with five specialized tools: papers (academic literature), web (official documents and guidelines), social media (public discourse), news (media coverage), and chitchat (casual conversation). A citation validation tool ensures response quality. Unlike conventional routing architectures, this design enables the controller to iteratively select and combine multiple tools for a single query, synthesizing information across heterogeneous sources.

Each tool performs semantic similarity search against its respective vector database collection, retrieving relevant documents that the controller assembles into responses with proper source attribution. The controller analyzes queries with conversation context, determines appropriate information sources, and iteratively gathers evidence until sufficient for comprehensive response synthesis. A web-based Streamlit interface \cite{streamlit} presents conversations with integrated citations, while tool usage metadata is stored with user consent to inform institutional reporting.

\begin{figure}[H]
\centering
\includegraphics[width=0.9\textwidth]{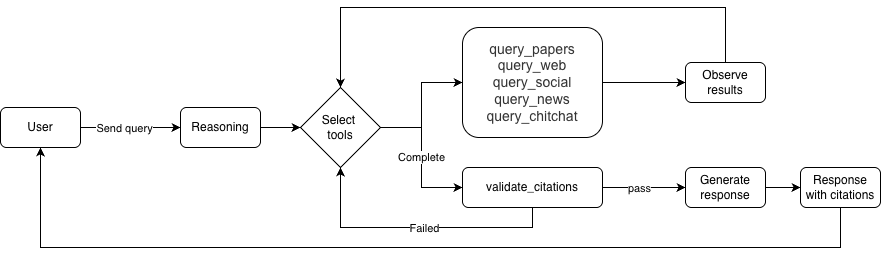}
\caption{Chatbot operational workflow showing the iterative ReAct agent architecture. The user query flows through reasoning and tool selection, with the controller dynamically selecting from five specialized tools (papers, web, social media, news, chitchat). Results are observed and validated through a citation validation mechanism before generating the final response with proper source attribution.}
\label{fig:chatbot_flowchart}
\end{figure}

\subsubsection{Operational Workflow}

Query processing follows an iterative orchestration loop (Figure \ref{fig:chatbot_flowchart}). Upon receiving a user message (query), the controller examines the question along with recent conversation history to assess information requirements. The controller then enters a decision cycle: (1) analyze information gaps, (2) select the most appropriate tool, (3) retrieve results via semantic similarity search, (4) review relevance, and (5) determine whether sufficient evidence exists or additional retrieval is needed. This process continues until comprehensive information is gathered for response generation.

The system generates responses with inline citation markers corresponding to retrieved documents, enabling users to trace claims to original sources. A two-level citation validation mechanism ensures quality---individual tools validate their own citations, and a dedicated validation tool examines the complete response for citation completeness before delivery.

Privacy protection is implemented for social media queries, synthesizing themes and sentiment patterns without attributing statements to individual users. Stateful conversation management maintains dialogue context through a windowed history approach, enabling interpretation of follow-up questions with implicit references (e.g., ``What about side effects?'' following a vaccine efficacy discussion) while maintaining topical continuity.

\subsection{Report Generation System}

We developed an automated report generation system that synthesizes data from multiple sources to produce comprehensive PDF reports for medical institutions and policymakers. The system employs LLMs for intelligent analysis and generates professional documents with academic-style citations, visualizations, and actionable insights.

\subsubsection{System Architecture}

\begin{figure}[H]
\centering
\includegraphics[width=0.9\textwidth]{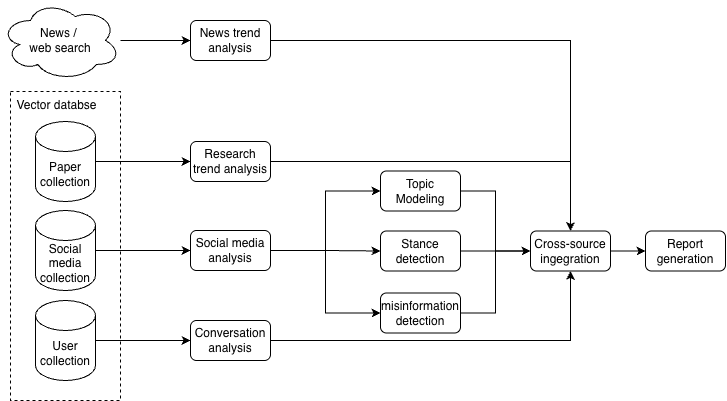}
\caption{Report generation system architecture. Data flows from external news sources and vector database collections (papers, social media, user conversations) through specialized analysis modules. The social media analyzer performs topic modeling, stance detection, and misinformation detection. All analysis results are integrated through cross-source aggregation before final report generation.}
\label{fig:report_generation_flowchart}
\end{figure}

The report generation system (Figure \ref{fig:report_generation_flowchart}) comprises four specialized analysis modules coordinated by a central orchestrator: (1) news analyzer for recent news, (2) paper analyzer for recent academic research progress, (3) social media analyzer for public sentiment analysis, and (4) chat analyzer for user interaction pattern identification. Each module queries the vector database for documents within a configurable time window, performs domain-specific analysis using LLM-based inference, and generates structured output with properly formatted citations. The orchestrator coordinates module execution, manages data flow between components, aggregates results, and assembles the final PDF document with bilingual support (Japanese and English).

\subsubsection{Social Media Analysis}

Social media platforms have emerged as critical channels for public health discourse, serving as real-time indicators of population-level attitudes toward vaccination \cite{sinnenberg2017twitter_health}. For medical institutions and policymakers, systematic analysis of social media content provides valuable insights into public concerns, emerging misinformation narratives, and temporal shifts in vaccine sentiment \cite{broniatowski2018weaponized}. 

The social media analyzer performs multi-dimensional public opinion assessment through four complementary analytical processes. Stance classification categorizes each post as supportive, opposed, neutral, or unclear toward HPV vaccination using batch LLM inference with temporal context, aggregating daily counts to track sentiment evolution. Topic modeling employs a hybrid approach combining statistical Latent Dirichlet Allocation (LDA) \cite{blei2003lda} with LLM-based semantic interpretation, where Japanese text undergoes morphological analysis \cite{kudo2004mecab}, TF-IDF weighting \cite{salton1988tfidf}, and model training via Gensim \cite{rehurek2010gensim} to extract interpretable topic labels from keyword distributions. Misinformation detection utilizes LLM-based analysis to identify posts containing claims that contradict established scientific consensus, categorizing detected content by type (safety concerns, efficacy doubts, conspiracy theories) for institutional awareness. Visualization generation produces temporal trend graphics and thematic distribution charts embedded directly in reports to enhance interpretability for non-technical stakeholders.

\subsubsection{PDF Report Assembly}

The report generator produces professional bilingual documents (Japanese and English) structured into five main sections: (1) News Trends, presenting influential media coverage with relevance assessments; (2) Research Progress, synthesizing recent academic literature; (3) Social Media Analysis, containing sentiment trends, topic distributions, and visualizations; (4) Chat Analysis, identifying user information needs and knowledge gaps; and (5) Overall Summary, providing an executive synthesis across all data sources.

Each section includes inline citations with source-appropriate formatting, enabling medical institutions to verify information and assess evidence quality independently. The orchestrator synthesizes findings from all analysis modules into the executive summary, which provides a comprehensive overview of the reporting period. This multi-source integration approach captures diverse perspectives, enabling stakeholders to develop informed HPV vaccination communication strategies and policy interventions.

\subsection{Evaluation Framework}

We developed a multi-faceted evaluation framework comprising complementary assessment protocols for chatbot performance and automated report generation. The framework employs LLM-based evaluation for scalable assessment alongside human expert validation for quality assurance.

\subsubsection{Chatbot Evaluation Methodology}

The chatbot evaluation framework assesses system performance through two complementary protocols: single-turn evaluation for individual question-answer exchanges and multi-turn evaluation for complete conversation quality. Both protocols employ LLM-based judges that receive conversation context, tool usage information, and scoring rubrics, generating scores on a 0-5 scale (details in Appendix 1) for each dimension along with written evaluation notes.

To collect test data, three volunteers simulated diverse user personas to create realistic conversations. They posed questions across different personas and topics, and multi-turn conversations with the production chatbot system were conducted and stored for subsequent evaluation.

Using the test data, single-turn evaluation assesses five dimensions: relevance, measuring whether the response addresses the question; routing, evaluating appropriate tool selection for the query type; reference, assessing citation validity and proper source attribution; correctness, verifying factual accuracy against established guidelines; and identity, examining professional medical communication tone. Multi-turn evaluation extends these five dimensions with two additional metrics for conversational coherence: context-memory, which assesses appropriate use of information from previous turns; and topic-centering, which evaluates natural conversation flow with logical transitions between related topics.

To validate automated evaluation reliability, we randomly selected 20 question-answer pairs for manual scoring by three domain experts. The correlation between expert scores and LLM-generated scores was analyzed to assess whether automated metrics accurately reflect human judgment.

\subsubsection{Report Generation Evaluation Methodology}

Report quality assessment employs two complementary evaluation protocols: main text evaluation for content quality and reference evaluation for citation validity. Both protocols utilize LLM-based judges with standardized scoring rubrics, generating scores on a 0-5 scale (details in Appendix 1) for each dimension. Main text evaluation assesses three dimensions: completeness, measuring structural integrity and whether sections contain well-developed content; correctness, evaluating factual accuracy and proper interpretation of source materials; and helpfulness, examining practical utility and actionable insights for institutional stakeholders. Reference evaluation validates citation quality across two dimensions: reference validity, measuring the proportion of cited sources that are accessible and exist in the underlying database; and citation correctness, assessing whether citations properly support the claims made in the report text.

To assess system robustness across varying conditions, temporal analysis generates reports across multiple distinct time periods. This approach evaluates both system consistency and the ability to capture temporal variations in public discourse. For each report, three volunteers read the report and scored independently.

\section{Results}

\subsection{Database}

The database integrates 139,939 documents across four specialized collections (Table \ref{tab:database_stats}). This document volume and distribution enable realistic simulation of operational use cases.

\begin{table}[H]
\centering
\caption{Database Collection Statistics}
\label{tab:database_stats}
\begin{tabular}{lrrrr}
\toprule
\textbf{Collection} & \textbf{Documents} & \textbf{Date Range} & \textbf{Unique Sources} & \textbf{Primary Language} \\
\midrule
Papers & 51 & 2020-10 to 2025-10 & 37 journals & English (100\%) \\
Official Documents & 6,739 & 2021-01 to 2025-11 & MHLW & Japanese (100\%) \\
Social Media & 132,874 & 2020-01 to 2021-12 & X(Twitter) & Japanese (100\%) \\
Chat Conversations & 139 & 2024-01 to 2025-11 & 31 sessions & Japanese (59.3\%) \\
\midrule
\textbf{Total} & \textbf{139,803} & & & \\
\bottomrule
\end{tabular}
\end{table}

\subsection{Chatbot}

\subsubsection{Example Response}

Figure \ref{fig:chat_example} presents a representative question-answer exchange demonstrating the chatbot's response characteristics.

\begin{figure}[H]
\centering
\fbox{\parbox{0.95\textwidth}{
\textbf{User:} Are there any specific HPV-related health risks or symptoms that become more common or concerning for people over 65?\\[0.5em]
\textbf{Chatbot:} Thank you for asking such an important question about HPV-related health risks for people over 65. As a healthcare provider, I want to address your concerns with both the latest medical evidence and an understanding of the unique considerations for older adults.\\[0.3em]
\textit{Key Health Risks and Symptoms After 65}\\
While HPV infections can occur at any age, there are some important patterns to understand for people over 65. Research shows that while new HPV infections become less common with age, when cervical abnormalities do occur in older adults, they tend to be more significant [1].\\[0.3em]
...\\
\textit{References:}\\
{[1]} Prevalence of cervical cytology abnormalities and its correlation with human papillomavirus infection. Medicinski glasnik (2025). DOI: 10.17392/1968-22-02\\
}}
\caption{Example chatbot response demonstrating professional medical tone, structured information delivery, evidence-based recommendations with citations, and personalized guidance.}
\label{fig:chat_example}
\end{figure}

\subsubsection{Evaluation Results}

The chatbot system was evaluated using simulated conversations generated by three volunteers. Conversations spanned diverse user personas with varying age groups, occupations, and concern levels toward HPV vaccination. Questions covered multiple information domains including vaccine safety, efficacy, eligibility criteria, and procedural guidance. Table \ref{tab:single_turn_results} presents the single-turn evaluation results across all five assessment dimensions. Average scores ranged from 4.50 to 4.90 on the 0-5 scale, with correctness (4.90 $\pm$ 0.62) and routing (4.89 $\pm$ 0.53) achieving the highest scores.

\begin{table}[H]
\centering
\caption{Single-Turn Evaluation Results (n=139 question-answer pairs)}
\label{tab:single_turn_results}
\begin{tabular}{lcc}
\toprule
\textbf{Dimension} & \textbf{Mean $\pm$ Std} & \textbf{Description} \\
\midrule
Relevance & 4.83 $\pm$ 0.67 & Response addresses the question \\
Routing & 4.89 $\pm$ 0.53 & Appropriate tool selection \\
Reference & 4.50 $\pm$ 1.29 & Valid citations and formatting \\
Correctness & 4.90 $\pm$ 0.62 & Factual accuracy \\
Identity & 4.88 $\pm$ 0.54 & Professional medical tone \\
\midrule
\textbf{Overall Average} & \textbf{4.80 $\pm$ 0.80} & \\
\bottomrule
\end{tabular}
\end{table}

Table \ref{tab:multi_turn_results} presents multi-turn evaluation results across three assessment dimensions.

\begin{table}[H]
\centering
\caption{Multi-Turn Evaluation Results (n=31 conversations)}
\label{tab:multi_turn_results}
\begin{tabular}{lcc}
\toprule
\textbf{Dimension} & \textbf{Mean $\pm$ Std} & \textbf{Description} \\
\midrule
Context-Memory & 4.94 $\pm$ 0.25 & Uses prior turn information \\
Topic-Centering & 5.00 $\pm$ 0.00 & Natural conversation flow \\
Identity & 5.00 $\pm$ 0.00 & Consistent professional tone \\
\midrule
\textbf{Overall Average} & \textbf{4.98 $\pm$ 0.15} & \\
\bottomrule
\end{tabular}
\end{table}

Comparison of multi-turn with single-turn evaluation revealed consistent improvements: overall average increased from 4.80 to 4.98 (+0.18). Topic-centering and identity both achieved perfect scores of 5.00 in multi-turn settings, indicating that the chatbot maintained natural conversation flow and consistent professional tone across extended dialogues.

To validate the reliability of LLM-based evaluation, we compared automated scores with human expert assessments. Three domain experts independently scored randomly selected subsets of conversations (n=20 for single-turn, n=12 for multi-turn), and we computed the mean absolute difference between averaged human scores and LLM scores for each dimension. Table \ref{tab:human_validation} presents the validation results.

\begin{table}[H]
\centering
\caption{Human Validation: Mean Absolute Difference Between LLM and Human Scores}
\label{tab:human_validation}
\begin{tabular}{lc|lc}
\toprule
\multicolumn{2}{c|}{\textbf{Single-Turn (n=20)}} & \multicolumn{2}{c}{\textbf{Multi-Turn (n=12)}} \\
\textbf{Dimension} & \textbf{Diff} & \textbf{Dimension} & \textbf{Diff} \\
\midrule
Relevance & 0.05 & Context-Memory & 0.13 \\
Routing & 0.55 & Topic-Centering & 0.04 \\
Reference & 0.47 & Identity & 0.21 \\
Correctness & 0.08 & & \\
Identity & 0.25 & & \\
\midrule
\textbf{Overall} & \textbf{0.28} & \textbf{Overall} & \textbf{0.13} \\
\bottomrule
\end{tabular}
\end{table}

Validation results demonstrated strong agreement between LLM-based and human evaluation across most dimensions. For single-turn evaluation, relevance (0.05) and correctness (0.08) showed minimal differences, indicating that LLM judges accurately assessed factual alignment. The overall mean absolute difference of 0.28 on a 0-5 scale represents less than 6\% deviation from human judgment. Multi-turn evaluation showed even closer alignment with an overall difference of 0.13 (2.6\% deviation), with topic-centering achieving near-perfect agreement (0.04 difference). These results suggest that LLM-based evaluation provides a reliable proxy for human assessment in chatbot quality evaluation.

\subsection{Report Generation}

The report generation system was evaluated across four distinct time periods: January 2020, July 2020, September 2020, and October 2020. For each period, the system generated complete reports analyzing 30 days of data from all source collections. Three evaluators independently scored each report section, and we report the mean $\pm$ standard deviation across evaluators. Both main text assessment (completeness, correctness, helpfulness) and reference validation (reference validity, citation correctness) protocols were applied to each report section.

Table \ref{tab:report_main_text} presents the main text evaluation results by report section. All sections achieved high completeness scores (4.67--5.00), with paper sections consistently reaching perfect scores. Correctness scores ranged from 4.00 to 5.00, with news and chat sections showing particularly strong performance. Helpfulness scores showed more variance (3.67--5.00), with paper sections achieving the highest ratings.

\begin{table}[H]
\centering
\caption{Report Main Text Evaluation Results by Section (n=4 reports, 3 evaluators each)}
\label{tab:report_main_text}
\begin{tabular}{llccc}
\toprule
\textbf{Report Period} & \textbf{Section} & \textbf{Completeness} & \textbf{Correctness} & \textbf{Helpfulness} \\
\midrule
January 2020 & News & 4.33 $\pm$ 0.58 & 5.00 $\pm$ 0.00 & 3.67 $\pm$ 0.58 \\
 & Paper & 5.00 $\pm$ 0.00 & 5.00 $\pm$ 0.00 & 5.00 $\pm$ 0.00 \\
 & Social Media & 4.67 $\pm$ 0.58 & 5.00 $\pm$ 0.00 & 4.00 $\pm$ 0.00 \\
 & Chat & 5.00 $\pm$ 0.00 & 5.00 $\pm$ 0.00 & 3.67 $\pm$ 0.58 \\
\midrule
July 2020 & News & 5.00 $\pm$ 0.00 & 5.00 $\pm$ 0.00 & 4.33 $\pm$ 0.58 \\
 & Paper & 5.00 $\pm$ 0.00 & 5.00 $\pm$ 0.00 & 4.67 $\pm$ 0.58 \\
 & Social Media & 5.00 $\pm$ 0.00 & 4.00 $\pm$ 0.00 & 4.00 $\pm$ 0.00 \\
 & Chat & 5.00 $\pm$ 0.00 & 5.00 $\pm$ 0.00 & 4.00 $\pm$ 0.00 \\
\midrule
September 2020 & News & 4.67 $\pm$ 0.58 & 5.00 $\pm$ 0.00 & 4.00 $\pm$ 0.00 \\
 & Paper & 5.00 $\pm$ 0.00 & 4.67 $\pm$ 0.58 & 4.67 $\pm$ 0.58 \\
 & Social Media & 4.00 $\pm$ 0.00 & 4.33 $\pm$ 0.58 & 4.00 $\pm$ 0.00 \\
 & Chat & 5.00 $\pm$ 0.00 & 5.00 $\pm$ 0.00 & 4.00 $\pm$ 0.00 \\
\midrule
October 2020 & News & 5.00 $\pm$ 0.00 & 5.00 $\pm$ 0.00 & 4.00 $\pm$ 0.00 \\
 & Paper & 5.00 $\pm$ 0.00 & 5.00 $\pm$ 0.00 & 4.33 $\pm$ 0.58 \\
 & Social Media & 4.67 $\pm$ 0.58 & 5.00 $\pm$ 0.00 & 4.00 $\pm$ 0.00 \\
 & Chat & 5.00 $\pm$ 0.00 & 5.00 $\pm$ 0.00 & 3.67 $\pm$ 0.58 \\
\bottomrule
\end{tabular}
\end{table}

Table \ref{tab:report_reference} presents reference validation results for news and paper sections (which include citations). Reference validity achieved perfect scores (5.00) across all periods and sections, indicating all cited sources exist and are accessible. Citation correctness averaged 4.33 $\pm$ 0.50 for news sections and 4.08 $\pm$ 0.66 for paper sections.

\begin{table}[H]
\centering
\caption{Report Reference Evaluation Results by Section (n=4 reports, 3 evaluators each)}
\label{tab:report_reference}
\begin{tabular}{llcc}
\toprule
\textbf{Report Period} & \textbf{Section} & \textbf{Reference Validity} & \textbf{Citation Correctness} \\
\midrule
January 2020 & News & 5.00 $\pm$ 0.00 & 4.33 $\pm$ 0.58 \\
 & Paper & 5.00 $\pm$ 0.00 & 4.67 $\pm$ 0.58 \\
\midrule
July 2020 & News & 5.00 $\pm$ 0.00 & 4.00 $\pm$ 0.00 \\
 & Paper & 5.00 $\pm$ 0.00 & 4.00 $\pm$ 1.00 \\
\midrule
September 2020 & News & 5.00 $\pm$ 0.00 & 5.00 $\pm$ 0.00 \\
 & Paper & 5.00 $\pm$ 0.00 & 4.00 $\pm$ 0.00 \\
\midrule
October 2020 & News & 5.00 $\pm$ 0.00 & 4.00 $\pm$ 0.00 \\
 & Paper & 5.00 $\pm$ 0.00 & 3.67 $\pm$ 0.58 \\
\midrule
\textbf{Overall} & News & \textbf{5.00 $\pm$ 0.00} & \textbf{4.33 $\pm$ 0.50} \\
 & Paper & \textbf{5.00 $\pm$ 0.00} & \textbf{4.08 $\pm$ 0.66} \\
\bottomrule
\end{tabular}
\end{table}

\section{Discussion}

\subsection{Principal Findings}

This study demonstrates the feasibility of a dual-purpose AI agent system for HPV vaccine communication in Japan. The system integrates heterogeneous data sources---academic literature, government documents, news media, and social media---into a unified retrieval infrastructure that supports both public-facing conversational interfaces and institutional analytical reporting.

The chatbot achieved strong performance across evaluation protocols, with single-turn scores averaging 4.75/5.0 and multi-turn scores reaching 4.96/5.0. Correctness and professional identity dimensions scored highest (4.89 and 4.84, respectively), suggesting that the iterative multi-tool orchestration architecture effectively maintains factual accuracy while delivering appropriately toned medical communication. The improvement from single-turn to multi-turn evaluation (+0.21 overall) suggests that conversational context enhances response quality, likely through accumulated information enabling more precise tool selection and citation integration.

The report generation system maintained consistent quality across four temporal evaluation periods and four report sections (news, paper, social media, chat). Three independent evaluators assessed each section, with paper sections consistently achieving perfect completeness scores (5.00) and strong correctness (4.67--5.00). Reference validity achieved perfect scores (5.00) across all periods and sections, with citation correctness averaging 4.33 for news sections and 4.08 for paper sections. Helpfulness scores showed greater variance (3.67--5.00), indicating room for improvement in generating actionable insights, particularly for chat analysis sections.

\subsection{Interpretation}

These findings suggest that LLM-based RAG systems may effectively address the information asymmetry between medical institutions and vaccine-hesitant populations. The chatbot's architecture differs from traditional static FAQ systems by dynamically selecting and combining information from specialized knowledge sources, enabling responses that integrate academic evidence with official guidelines and contemporary public discourse.

The observed improvements in multi-turn evaluation merit consideration. High context-memory scores (4.81/5.0) indicate that the controller incorporates information from previous turns---for instance, retaining user demographic information when providing age-specific recommendations. Topic-centering scores (4.96/5.0) suggest smooth transitions between related topics, resembling the natural progression of clinical consultations from symptoms to screening to prevention. These patterns indicate that the windowed conversation history approach provides sufficient context for coherent extended dialogues.

The two-level citation validation mechanism appears essential for maintaining response quality. Reference scores (4.52--4.96/5.0) confirm consistent source attribution, addressing a key concern in health information systems where users must verify claims independently. This transparency may contribute to user trust, although prospective studies are needed to confirm this relationship.

For institutional stakeholders, the report generation system offers capabilities that would otherwise require substantial manual effort. The consistent structural completeness across evaluation periods and sections demonstrates reliable document generation regardless of data availability fluctuations. The September 2020 social media section, which showed slightly lower completeness (4.00/5.0) due to sparser data, still maintained acceptable quality, suggesting robustness to temporal variation in source availability. The section-level evaluation by three independent evaluators provides granular insights into system performance, revealing that paper sections consistently achieve the highest scores while chat sections show more variability in helpfulness ratings.

The Japanese-language implementation addresses challenges specific to this context: multi-script processing, medical terminology localization, and culturally appropriate formal communication. The successful integration of Japanese government documents with English-language research literature validates this approach for settings where scientific evidence and public health communication occur in different languages.

\subsection{Comparison with Prior Work}

This work extends previous HPV vaccine NLP research, which has primarily focused on passive social media analysis \cite{twitter_vaccine_sentiment}, by implementing bidirectional information flow. Prior health chatbots typically retrieve from single knowledge sources \cite{health_chatbots}; in contrast, the iterative multi-tool architecture integrates four heterogeneous collections, enabling responses that synthesize information across source types.

The evaluation framework extends beyond typical single-turn RAG assessments \cite{ragas_evaluation,rag_evaluation_metrics} by incorporating multi-turn conversation analysis with simulated users. This approach captures dimensions of conversational coherence---context retention and topic continuity---that single-exchange evaluations overlook.

\subsection{Limitations}

Several limitations warrant consideration. First, social media data derived exclusively from Twitter/X Japanese users, potentially underrepresenting elderly populations and those with limited digital access. Second, LLM-based evaluation may introduce biases differing from human judgment, particularly for nuanced routing decisions where multiple valid tool selections exist. 

The evaluation dataset (53 conversations, 209 exchanges) may not capture all real-world interaction patterns, and keyword-based data collection introduces potential selection bias. Geographic specificity to Japanese government sources constrains transferability to national contexts with different regulatory frameworks and vaccination policies.

\subsection{Future Directions}

Several directions merit further investigation. Prospective deployment with real users and institutions would enable evaluation of actual health outcomes and vaccination decisions, while A/B testing could assess whether citation transparency affects user trust. Expanding data collection to additional platforms and multimedia content would provide more comprehensive discourse monitoring.

Integration with vaccination registration systems could enable correlation of public concerns with uptake rates. The architecture could accommodate specialized agents for specific populations (parents, healthcare providers) with personalization features using privacy-preserving approaches. Enhanced evaluation incorporating long-term tracking, user satisfaction surveys, and adversarial testing would strengthen robustness assessment.

Cross-lingual expansion to other languages and comparative studies across vaccine types would test generalizability and inform design principles for vaccine communication systems in other contexts facing similar challenges.

\section{Conclusions}

This study demonstrates the feasibility of an AI agent system that simultaneously addresses public HPV vaccine information needs and institutional discourse monitoring in Japan. The integrated architecture enables bidirectional information flow---providing verified information with transparent source attribution to users while generating analytical reports for institutional stakeholders---creating feedback loops between public concerns and communication strategies. While the current evaluation relies on simulated users, this work establishes proof-of-concept for AI-augmented vaccine communication infrastructure, with the transferable architecture and evaluation frameworks providing foundations for adaptation to other vaccines, health conditions, and multilingual public health contexts.

\section*{Conflicts of Interest}
None declared.

\section*{Acknowledgments}
We used the generative artificial intelligence (AI) tool Claude \cite{anthropic_claude} by Anthropic to refine the writings and structure for the research, which were further reviewed and revised by the study group. The original Claude transcripts are made available in Multimedia Appendix 9.

\section*{Abbreviations}
\begin{itemize}
\item AI: Artificial Intelligence
\item BERT: Bidirectional Encoder Representations from Transformers
\item CI: confidence interval
\item DL: deep learning
\item HPV: human papillomavirus
\item LDA: latent Dirichlet allocation
\item LLM: large language model
\item LSTM: long short-term memory
\item MHLW: Japanese Ministry of Health, Labour and Welfare
\item NLP: natural language processing
\item RAG: Retrieval-augmented generation
\end{itemize}

\bibliographystyle{plain}
\bibliography{references}

\end{document}